\documentclass{bmvc2k}
\pdfoutput=1

\title{A Study on Unsupervised Domain Adaptation for Semantic Segmentation in the Era of Vision-Language Models}

\addauthor{Manuel Schwonberg}{schwonberg@campus.tu-berlin.de}{*1,3}
\addauthor{Claus Werner}{claus.werner@cariad.technology}{*2,3}
\addauthor{Hanno  Gottschalk}{gottschalk@math.tu-berlin.de}{1}
\addauthor{Carsten Meyer}{carsten.meyer@ostfalia.de}{2,4}
\addauthor{\tiny \vspace{-0.5cm} * equal contribution}{}{}

\addinstitution{
 Mathematical Modeling of Industrial Life Cycles, TU Berlin\\
}
\addinstitution{
 Department of Computer Science,\\
 Ostfalia University of Applied Sciences\\
}
\addinstitution{
 CARIAD SE\\
}
\addinstitution{
 Department of Computer Science,\\
 Kiel University \\
}

\runninghead{Schwonberg, Werner \etal}{Study on Unsupervised Domain Adaptation}

\def\etal{\emph{et al}\bmvaOneDot}

\newcommand{\src}[0]{{\mathcal{D}^\mathrm{S}}}
\newcommand{\tgt}[0]{{\mathcal{D}^\mathrm{T}}}

\newcommand{\cstrain}[0]{{\mathcal{D}^\mathrm{CS}_\mathrm{train}}}
\newcommand{\csval}[0]{{\mathcal{D}^\mathrm{CS}_\mathrm{val}}}

\newcommand{\bddval}[0]{{\mathcal{D}^\mathrm{BDD}_\mathrm{val}}}

\newcommand{\mvval}[0]{{\mathcal{D}^\mathrm{MV}_\mathrm{val}}}

\newcommand{\acdctrain}[0]{{\mathcal{D}^\mathrm{ACDC}_\mathrm{train}}}
\newcommand{\acdcval}[0]{{\mathcal{D}^\mathrm{ACDC}_\mathrm{val}}}
\newcommand{\acdctest}[0]{{\mathcal{D}^\mathrm{ACDC}_\mathrm{test}}}

\newcommand{\acdcnormal}[0]{{\mathcal{D}^\mathrm{ACDC}_\mathrm{normal}}}
\newcommand{\acdcfog}[0]{{\mathcal{D}^\mathrm{ACDC}_\mathrm{fog}}}
\newcommand{\acdcsnow}[0]{{\mathcal{D}^\mathrm{ACDC}_\mathrm{snow}}}
\newcommand{\acdcrain}[0]{{\mathcal{D}^\mathrm{ACDC}_\mathrm{rain}}}
\newcommand{\acdcnight}[0]{{\mathcal{D}^\mathrm{ACDC}_\mathrm{night}}}

\newcommand{\gtavfull}[0]{{\mathcal{D}^\mathrm{GTA5}}}

\newcommand{\putindex}[3]{\vtop{\hbox{\hspace{#3} $#1$}
            \hbox{\raise 6mm \hbox{$\scriptscriptstyle #2$}}}}

\newcommand{\gradx}[0]{\vtop{\hbox{\rm grad}
            \hbox{\raise 2.5mm \hbox{\rm \hspace{2mm} \footnotesize x}}}}

\newcommand{\grady}[0]{\vtop{\hbox{\rm grad}
            \hbox{\raise 2.5mm \hbox{\rm \hspace{2mm} \footnotesize y}}}}

\newcommand{\grad}[1]{\vtop{\hbox{\rm grad}
            \hbox{\raise 2.5mm \hbox{#1}}}}

\newcommand{\btb}{     \begin{tabbing}             }
\newcommand{\bte}{     \end{tabbing}               }
\definecolor{tu0}{rgb}{0.7451, 0.1176, 0.2353}

\definecolor{tu1}{rgb}{1.0000, 0.8039, 0.0000}
\definecolor{tu11}{rgb}{1.0000, 0.8627, 0.3020}
\definecolor{tu12}{rgb}{1.0000, 0.9020, 0.4980}
\definecolor{tu13}{rgb}{1.0000, 0.9412, 0.6980}
\definecolor{tu14}{rgb}{1.0000, 0.9608, 0.8000}

\definecolor{tu2}{rgb}{0.9804, 0.4314, 0.0000}
\definecolor{tu21}{rgb}{0.9882, 0.6039, 0.3020}
\definecolor{tu22}{rgb}{0.9882, 0.7137, 0.4980}
\definecolor{tu23}{rgb}{0.9922, 0.8275, 0.6980}
\definecolor{tu24}{rgb}{0.9961, 0.8863, 0.8000}

\definecolor{tu3}{rgb}{0.6902, 0.0000, 0.2745}
\definecolor{tu31}{rgb}{0.7529, 0.2000, 0.4196}
\definecolor{tu32}{rgb}{0.8431, 0.4980, 0.6353}
\definecolor{tu33}{rgb}{0.9216, 0.7490, 0.8196}
\definecolor{tu34}{rgb}{0.9529, 0.8510, 0.8902}

\definecolor{tu4}{rgb}{0.4863, 0.8039, 0.9020}
\definecolor{tu41}{rgb}{0.6431, 0.8627, 0.9333}
\definecolor{tu42}{rgb}{0.7412, 0.9020, 0.9490}
\definecolor{tu43}{rgb}{0.8431, 0.9412, 0.9686}
\definecolor{tu44}{rgb}{0.8980, 0.9608, 0.9804}

\definecolor{tu5}{rgb}{0.0000, 0.5020, 0.7059}
\definecolor{tu51}{rgb}{0.3020, 0.6510, 0.7961}
\definecolor{tu52}{rgb}{0.5490, 0.7765, 0.8667}
\definecolor{tu53}{rgb}{0.7490, 0.8745, 0.9255}
\definecolor{tu54}{rgb}{0.8510, 0.9255, 0.9569}

\definecolor{tu6}{rgb}{0.0000, 0.3255, 0.4549}
\definecolor{tu61}{rgb}{0.2510, 0.4941, 0.5922}
\definecolor{tu62}{rgb}{0.5490, 0.6941, 0.7529}
\definecolor{tu63}{rgb}{0.7490, 0.8314, 0.8627}
\definecolor{tu64}{rgb}{0.8510, 0.8980, 0.9176}

\definecolor{tu7}{rgb}{0.0314, 0.0314, 0.0314}
\definecolor{tu71}{rgb}{0.3725, 0.3725, 0.3725}
\definecolor{tu72}{rgb}{0.5882, 0.5882, 0.5882}
\definecolor{tu73}{rgb}{0.7529, 0.7529, 0.7529}
\definecolor{tu74}{rgb}{0.8667, 0.8667, 0.8667}

\definecolor{tu8}{rgb}{0.7765, 0.9333, 0.0000}
\definecolor{tu81}{rgb}{0.8431, 0.9529, 0.3020}
\definecolor{tu82}{rgb}{0.8863, 0.9647, 0.4980}
\definecolor{tu83}{rgb}{0.9333, 0.9804, 0.6980}
\definecolor{tu84}{rgb}{0.9569, 0.9882, 0.8000}

\definecolor{tu9}{rgb}{0.5373, 0.6431, 0.0000}
\definecolor{tu91}{rgb}{0.6784, 0.7490, 0.3020}
\definecolor{tu92}{rgb}{0.7686, 0.8196, 0.4980}
\definecolor{tu93}{rgb}{0.8588, 0.8941, 0.6980}
\definecolor{tu94}{rgb}{0.9059, 0.9294, 0.8000}

\definecolor{tu10}{rgb}{0.0000, 0.4431, 0.3373}
\definecolor{tu101}{rgb}{0.3020, 0.6118, 0.5373}
\definecolor{tu102}{rgb}{0.5490, 0.7490, 0.7020}
\definecolor{tu103}{rgb}{0.7490, 0.8588, 0.8353}
\definecolor{tu104}{rgb}{0.8549, 0.9176, 0.9059}

\definecolor{tu110}{rgb}{0.8000, 0.0000, 0.6000}
\definecolor{tu111}{rgb}{0.8706, 0.3490, 0.7412}
\definecolor{tu112}{rgb}{0.9216, 0.6000, 0.8392}
\definecolor{tu113}{rgb}{0.9608, 0.8000, 0.9216}
\definecolor{tu114}{rgb}{0.9804, 0.8980, 0.9608}

\definecolor{tu120}{rgb}{0.4627, 0.0000, 0.4627}
\definecolor{tu121}{rgb}{0.5961, 0.2510, 0.5961}
\definecolor{tu122}{rgb}{0.7294, 0.4980, 0.7294}
\definecolor{tu123}{rgb}{0.8392, 0.6980, 0.8392}
\definecolor{tu124}{rgb}{0.9216, 0.8510, 0.9216}

\definecolor{tu130}{rgb}{0.4627, 0.0000, 0.3294}
\definecolor{tu131}{rgb}{0.6118, 0.3020, 0.5333}
\definecolor{tu132}{rgb}{0.7569, 0.5490, 0.6980}
\definecolor{tu133}{rgb}{0.8667, 0.7490, 0.8314}
\definecolor{tu134}{rgb}{0.9216, 0.8510, 0.9020}

\definecolor{con_blue}{HTML}{004488}
\definecolor{con_yellow}{HTML}{DDAA33}
\definecolor{con_red}{HTML}{BB5566}
\definecolor{con_black}{HTML}{000000}
\definecolor{con_gray}{HTML}{AAAAAA}

\definecolor{vib_orange}{HTML}{EE7733}
\definecolor{vib_darkblue}{HTML}{0077BB}
\definecolor{vib_lightblue}{HTML}{33BBEE}
\definecolor{vib_magenta}{HTML}{EE3377}
\definecolor{vib_red}{HTML}{CC3311}
\definecolor{vib_green}{HTML}{009988}
\definecolor{vib_gray}{HTML}{BBBBBB}
\definecolor{vib_black}{HTML}{000000}
\usepackage{booktabs}
\usepackage{multirow} 
\usepackage{cleveref}
\usepackage{tikz}
\usepackage{pgfplots}\pgfplotsset{compat=1.9}
\usepackage{wrapfig}
\usepackage{floatflt}
\usepackage{colortbl}
\usepackage{capt-of}
\hypersetup{colorlinks=true}

\definecolor{textcolor}{RGB}{70,170,34}
\begin{document}

\maketitle

\begin{abstract}
Despite the recent progress in deep learning based computer vision, domain shifts are still one of the major challenges. Semantic segmentation for autonomous driving faces a wide range of domain shifts, e.g. caused by changing weather conditions, new geolocations and the frequent use of synthetic data in model training. Unsupervised domain adaptation (UDA) methods have emerged which adapt a model to a new target domain by only using unlabeled data of that domain. The variety of UDA methods is large but all of them use ImageNet pre-trained models. 
Recently, vision-language models have demonstrated strong generalization capabilities which may facilitate domain adaptation. We show that simply replacing the encoder of existing UDA methods like DACS by a vision-language pre-trained encoder can result in significant performance improvements of up to 10.0\% mIoU on the GTA5$\rightarrow$Cityscapes domain shift. For the generalization performance to unseen domains, the newly employed vision-language pre-trained encoder provides a gain of up to 13.7\% mIoU across three unseen datasets. However, we find that not all UDA methods can be easily paired with the new encoder and that the UDA performance does not always likewise transfer into generalization performance. Finally, we perform our experiments on an adverse weather condition domain shift to further verify our findings on a pure real-to-real domain shift.

\end{abstract}

\section{Introduction}
\label{sec:intro}
Computer vision has experienced several breakthroughs in the past decade enabled by deep neural networks (DNNs)~\cite{radford2021learning, he2016deep, krizhevsky2012imagenet, Dosovitskiy2021}. Domain shifts, i.e. when the training and inference distribution differ, are still a major challenge for DNNs and can cause severe performance drops, e.g. when models are trained on synthetic data and inference is done on real data~\cite{Hoyer2022daformer, Schwonberg2023Survey, tsai2018learning}. These can significantly hamper the application especially to safety-critical areas like autonomous driving and medical image analysis.
Consequently, the mitigation of domain shifts is a major objective and several research fields emerged.%

The most popular field is unsupervised domain adaptation (UDA) where only unlabeled samples from the target domain are available and the objective is to adapt the model towards this specific target domain. A broad variety of UDA methods have been developed in the past years utilizing methods like adversarial adaptation~\cite{tsai2018learning, Tsai2019}, contrastive learning~\cite{xie2023sepico, zhang2021prototypical, jiang2022prototypical}, self-training~\cite{tranheden2021dacs, zou2018unsupervised} and knowledge distillation frameworks~\cite{Hoyer2022daformer, zhang2021multiple, shen2023diga}. Next to unsupervised domain adaptation so-called domain generalization (DG) methods gained increasing research attention~\cite{niemeijer2024generalization, termohlen2023re, huang2021fsdr, 2023iccv_PASTA, kim2021wedge, Lee2022wildnet, Choi2019, zhong2022adversarial, yue2019domain, fahes2023poda, schwonberg2023augmentation}. Here, no target data is available and the objective is to obtain a model which generalizes well across multiple unseen target domains. Very recently, the utilization of vision-language models (VLMs) enabled a major performance increase as well as methodological progress in the field of domain generalization with works like Rein~\cite{wei2024stronger}, CLOUDS~\cite{benigmim2024collaborating}, and VLTSeg~\cite{hummer2023vltseg}. These approaches demonstrate that large-scale vision-language pre-training like CLIP~\cite{radford2021learning} 
can be leveraged to significantly improve domain generalized semantic segmentation as well as detection~\cite{vidit2023clip} performance. Surprisingly, the field of UDA research falls back behind DG research because so far all existing UDA methods use ImageNet pre-trained models leaving the strong potential of vision-language models unexplored except a very recent study from Englert \etal~\cite{englert2024exploring}. For this reason, we equip selected UDA methods with a state-of-the-art vision-language pre-trained encoder and show their strong potential for unsupervised domain adaptation. We also evaluate the domain generalization capabilities of the UDA methods on unseen datasets and show that a simple UDA method with a vision-language pre-trained backbone provides state-of-the-art generalization capabilities. Next to the established synthetic-to-real and real-to-real benchmarks which contain a mixture of different domain shifts (e.g. GTA5$\rightarrow$Cityscapes contains synthetic and geolocation shift) we evaluate both the adaptation and generalization performance for a single pure adverse weather condition shift on the ACDC~\cite{sakaridis2021acdc} dataset. This is motivated by the results from Sakaridis \etal~\cite{sakaridis2021acdc} that UDA methods perform very different on different domain shifts and in some cases worsen the performance.
Overall, our study makes the following contributions and novel findings:

\begin{itemize}
    \item Extensive evaluation of UDA methods, in particular DACS \cite{tranheden2021dacs},  equipped with a state-of-the-art vision-language pre-trained backbone demonstrating that this can boost the UDA target performance by 10.0\% mIoU and across three unseen datasets by 13.7\% mIoU
     \item Analysis of UDA methods revealing that not all methods are similarly compatible with a
     VLM-based encoder; this indicates the need for new UDA methods tailored towards vision-language models 
     \item Extensive evaluation on unseen datasets showing that the UDA and generalization performance are not necessarily correlated and that recent DG methods can provide better generalization than UDA methods
\end{itemize}

\section{Related Work}
\textbf{Unsupervised Domain Adaptation (UDA) Methods} There exists a broad variety of UDA methods which can be coarsely clustered into input, feature, output space and hybrid adaptation methods~\cite{Schwonberg2023Survey}. In the input space several approaches apply a GAN-based style transfer between domains~\cite{chen2019crdoco, hoffman2018cycada, wang2020differential, Lee2019d, chang2019all} or augmentation and image mixing techniques~\cite{araslanov2021self, tranheden2021dacs,melas2021pixmatch}. In the feature space adversarial and contrastive learning UDA methods are often used~\cite{tsai2018learning, Hoffman2016, Luo2019c, marsden2022contrastive, liu2021domain}, whereas in the output space self-training, contrastive learning and knowledge distillation are common techniques~\cite{xie2023sepico, zou2018unsupervised, tsai2018learning, zhang2021prototypical}. Hybrid approaches combine multiple of the mentioned techniques and have become increasingly popular for UDA in recent years~\cite{Schwonberg2023Survey, wang2020differential}. However, hybrid approaches are mostly complex and for this reason we also include simpler approaches like DACS~\cite{tranheden2021dacs} in our study.\\       
\textbf{UDA Architectures} The vast majority of UDA approaches employed ImageNet pre-trained VGG-16~\cite{simonyan2014very} and ResNet-101~\cite{he2016deep} networks as their backbones~\cite{tsai2018learning, hoffman2018cycada, sankaranarayanan2018learning, lian2019constructing}, so that these architectures became the de-facto standard. With the emerging vision transformers and the foundational work DAFormer~\cite{Hoyer2022daformer} the mix vision transformer (MiT-B5) proposed by Xie \etal ~\cite{Xie2022segformer} became a popular backbone for UDA methods~\cite{hoyer2022hrda, xie2023sepico}. However, all of these backbones are ImageNet pre-trained and do not harness the strong generalization power of vision-language pre-training. Only the recent study from Englert \etal~\cite{englert2024exploring} investigates foundation models for UDA methods but our study differs in three important aspects. First, we also include previous, simpler UDA methods like DACS~\cite{tranheden2021dacs} and demonstrate that these methods also significantly benefit from vision-language pre-training. Second, Englert \etal~\cite{englert2024exploring} focus on the DINOv2 model~\cite{oquab2023dinov2} where the target domain Cityscapes~\cite{cordts2016cityscapes} is used to retrieve similar samples from the web and therefore results may not transfer to other settings. We instead focus on the vision-language pre-trained model EVA02-CLIP~\cite{EVA-CLIP}. Third, we follow the benchmark protocol from DG approaches~\cite{niemeijer2024generalization, 2023iccv_PASTA, wei2024stronger, huang2021fsdr, hummer2023vltseg} enabling a comparison to those studies. 
Our study shares similarities with the work from Piva \etal~\cite{piva2023empirical} w.r.t. their evaluation methodology; however, vision-language models are missing in their study and UDA and DG have made significant progress since then.\\
\textbf{UDA Benchmarks} Synthetic-to-real and real-to-real shifts are mostly used for benchmarking in UDA. The most common synthetic datasets are GTA5~\cite{richter2016playing} and SYNTHIA~\cite{ros2016synthia} and recently Urbansyn~\cite{gomez2023all}. As target domain for the synthetic-to-real domain shift the Cityscapes dataset~\cite{cordts2016cityscapes} is widely employed which is also often used as real source domain for the real-to-real domain shift. Real target domains are often the ACDC~\cite{sakaridis2021acdc}, FoggyCityscapes~\cite{sakaridis2018semantic} and DarkZurich~\cite{SDV19} datasets. All these domain shifts represent a mixture of at least two domain shifts or rely on artificially generated shifts like in the FoggyCityscapes dataset.\\
\textbf{Vision-Language Models (VLMs)} CLIP by Radford \etal~\cite{radford2021learning} was the foundational work in the field of vision-language models which was trained with image-text pairs and a contrastive loss for the alignment of vision and text embeddings. VLMs like CLIP benefit from large-scale multi-modal datasets like Laion-5B~\cite{schuhmann2022laionb} or Commonpool~\cite{gadre2024datacomp} and their size is an inherent advantage since the image-text pairs are easier to collect than e.g. single-class labels for ImageNet~\cite{radford2021learning}. 
VLMs are commonly used for pre-training and then employed in a transfer learning setting for a downstream task, e.g.\ semantic segmentation~\cite{luddecke2022image, ding2022decoupling, zhou2022extract, rao2022denseclip, ghiasi2022scaling} but not for UDA in semantic segmentation except the study from Englert \etal \cite{englert2024exploring}.

\section{Method}
In this section, we describe the details of our study, focusing on the %
evaluated UDA methods, the model architectures and the domain shifts of the ACDC dataset.
\subsection{UDA Methods}
In contrast to Englert \etal~\cite{englert2024exploring} we decide to not only focus on the most recent and usually more complex UDA methods, but deliberately also include previous methods. That shows how previous UDA methods and principles like adversarial adaptation
benefit from the new vision-language pre-trained backbone and how they perform on a pure domain shift.
We select previous, highly influential UDA methods: AdaptSegNet~\cite{tsai2018learning}, ADVENT~\cite{Vu2019} and DACS~\cite{tranheden2021dacs}.  In addition, we include the recent state-of-the-art methods SePiCo~\cite{xie2023sepico}, DAFormer~\cite{Hoyer2022daformer}, and MIC~\cite{hoyer2023mic}.
AdaptSegNet~\cite{tsai2018learning} is one of the earliest UDA works and utilizes adversarial domain adaptation by employing a domain discriminator in both the feature and output space. Similar to AdaptSegNet, ADVENT~\cite{Vu2019} applies adversarial learning and self-training on the entropy maps of the output space. DACS~\cite{tranheden2021dacs} is an easy-to-apply method which is incorporated by several subsequent UDA methods, combining input space cross domain image mixing and adaptive self-training.
SePiCo~\cite{xie2023sepico} as one of the current state-of-the-art methods proposes multiple contrastive losses along with a teacher-student framework to align the source and target domains. DAFormer~\cite{Hoyer2022daformer} was the first work using a vision transformer backbone for UDA and applied self-training, rare class sampling and an ImageNet feature distance loss to preserve ImageNet knowledge. We include both the initial DAFormer method and its follow-up approach MIC~\cite{hoyer2023mic} with HRDA \cite{hoyer2022hrda}. 
\subsection{UDA Architectures}
\textbf{Encoder \& Initialization} For the encoder choice, we follow recent domain generalization approaches~\cite{hummer2023vltseg, wei2024stronger} and employ the EVA02-CLIP-L-14 vision encoder~\cite{EVA-CLIP} which has shown strong generalization capabilities for segmentation. EVA02-CLIP~\cite{EVA-CLIP} relies on a sequence of CLIP and masked image modeling pre-training. Note that we only use the vision encoder of the pre-training and refer to it as EVA02-L. Hümmer \etal~\cite{hummer2023vltseg} demonstrated the strong generalization capabilities of the EVA02-CLIP encoder which makes it a natural candidate for our study. Both Wei \etal~\cite{wei2024stronger} and Englert \etal~\cite{englert2024exploring} focused on DINOv2~\cite{oquab2023dinov2} pre-trained weights as their initialization. We are not using DINOv2 pre-training in our study since the Cityscapes dataset, which is one of our main target domains, is used to sample the pre-training dataset of DINOv2, reducing the significance of evaluations.
Moreover, we include the established UDA architectures DeepLabv2 with a ResNet-101 backbone~\cite{chen2017deeplab} and the DAFormer architecture with a MiT-B5 backbone~\cite{Hoyer2022daformer} as it is common practice in UDA and DG benchmarking~\cite{Hoyer2022daformer, xie2023sepico, zhang2021prototypical, hoyer2023domain, niemeijer2024generalization}.\\ 
\textbf{Decoder} We employ an ASPP-based decoder with different dilation rates from the DAFormer architecture~\cite{Hoyer2022daformer}. The ASPP-decoder receives multi-level features from different levels of the encoder and performs up-sampling to obtain a common size of the feature maps
in case of a hierarchical encoder like a ResNet-101 or a MiT-B5. When using the EVA02-L encoder this upsamling has no effect since the encoder is non-hierarchical.

\subsection{Domain Shift Datasets}
Most of the real-to-real domain shifts for benchmarking are a mixture of at least two different domain shifts like Cityscapes$\rightarrow$ACDC and Cityscapes$\rightarrow$DarkZurich. Those benchmarks contain a geolocation shift, a weather/condition shift and also have been recorded with different sensor setups. In contrast, we aim to evaluate UDA methods in scenarios which exclusively cover only a single, well defined domain shift, e.g.\ only a geolocation or only an adverse weather condition shift. The available datasets for such an evaluation are limited. To the best of our knowledge only the ACDC~\cite{sakaridis2021acdc} and the DarkZurich~\cite{SDV19} datasets offer a well-defined single domain shift with 1:1 scene correspondences. DarkZurich is not included in our experiments because the pure day-to-nighttime shift is already contained in ACDC. Diverse datasets like BDD100K~\cite{yu2020bdd100k} or IDD~\cite{varma2019idd} do not contain the required metadata which enable a pure domain shift evaluation. Since ACDC offers direct scene correspondences between normal daytime weather and night, snow, fog and rain conditions, we chose the clean $\rightarrow$ adverse weather condition domain shift and the ACDC dataset for evaluation. We refer to the normal weather daytime images as $\acdcnormal$  and the adverse weather domains as $\acdcsnow$, $\acdcfog$ etc. while the official train, validation and test set with all subdomains are denoted as $\acdctrain$, $\acdcval$ and $\acdctest$. 

Next to this pure shift of the ACDC dataset we follow common practice in the UDA and DG field~\cite{Hoyer2022daformer, niemeijer2024generalization, wei2024stronger, fahes2024simple, tsai2018learning} and employ the GTA5~\cite{richter2016playing} and the SYNTHIA~\cite{ros2016synthia}(SYN) dataset as the synthetic source domains with 24966 and 9400 images, respectively. As the real-world domain we utilize Cityscapes~\cite{cordts2016cityscapes} (CS), Mapillary Vistas~\cite{neuhold2017mapillary} and BDD100K~\cite{yu2020bdd100k} with 2975/500, 18000/2000, 7000/1000 train/validation images respectively. We denote the respective datasets with a subscript as e.g. $\cstrain$ or $\csval$. All values of this study are reported on the respective validation datasets. For the DG evaluation only the validation sets of the corresponding domains are used. The ACDC dataset~\cite{sakaridis2021acdc} which is both used as source and target domain contains 1000 images in each sub-domain from which 400 are training, 100 validation and 500 are test images. For half of them reference images under normal weather conditions are available and were used for our new pure ACDC domain shift evaluation.
\subsection{Experimental Settings}
\textbf{Implementation Details} All experiments are based on the open source framework MMSegmentation~\cite{contributors2020mmsegmentation} and were conducted on a single A100 GPU with 80GB memory. The crop resolution for all experiments was fixed to 512$\times$512 except for MIC~\cite{hoyer2023mic}, where a 1024$\times$1024 resolution was used. The number of training iterations was set to 40k as common practice and a batch size of four was used. For ADVENT and AdaptSegNet with a ResNet-101 backbone the SGD optimizer with a learning rate of $2.5e-03$ was used. In all other cases, the AdamW~\cite{loshchilov2018decoupled} optimizer was selected in line with previous approaches~\cite{wei2024stronger, Hoyer2022daformer, hummer2023vltseg}. For the MiT-B5 backbone a learning rate of $6e-05$ and for the EVA02-L encoder of $1e-05$ was used. Hyperparameters specific to the respective UDA methods were set as given by the authors without change.\\
\textbf{Metric} As the evaluation metric we use the mean intersection over union (mIoU) averaged across 19 classes which are shared among all synthetic and real datasets. Only for SYNTHIA the mIoU across 13 classes is reported as common standard in UDA~\cite{xie2023sepico, Vu2019, shen2023diga, zhang2021prototypical}.   
\section{Results}
In this section, we show the results and start with the evaluation of the UDA performance with vision-language pre-training followed by the domain generalization evaluation. All results obtained with the vision-language pre-trained encoder EVA-02-CLIP will be highlighted with this \colorbox{textcolor!30}{green} color.  
\subsection{UDA with vision-language pre-training}
We equipped four of the selected UDA methods with the vision-language pre-trained EVA02-L encoder and compared it to the current two standard architectures ResNet-101 and MiT-B5, both initialized with ImageNet pre-trained weights. We could not include the combination of MIC and EVA02-CLIP in our study, but future work should investigate this combination. 
\begin{table}[ht!]
\centering
  
  \resizebox{\columnwidth}{!}{
  \begin{tabular}{ll|c|cc|cc|cc|c|cc}
  \toprule
  \multicolumn{12}{c}{\textbf{GTA5}$\rightarrow$\textbf{Cityscapes}}\\[0.5ex]
\toprule
       \multicolumn{2}{c|}{\textbf{Architecture}}     &       \multicolumn{8}{c|}{\textbf{UDA Method (in \% mIoU)}} &    &\\[0.5ex] %
   \textbf{Encoder} & \textbf{Decoder}       & \multicolumn{2}{c}{\textbf{AdaptSegNet~\cite{tsai2018learning}}}    &  \multicolumn{2}{c}{\textbf{ADVENT~\cite{Vu2019}}}  & \multicolumn{2}{c}{\textbf{DACS~\cite{tranheden2021dacs} }}     & \multicolumn{2}{c|}{\textbf{DAFormer~\cite{Hoyer2022daformer} }}   &\textbf{Src. Only} & \textbf{Oracle} \\
    \midrule
    && Abs. & Rel. &Abs. & Rel. & Abs. & Rel. &Abs. & Rel. && \\
    \cmidrule(lr){3-4} \cmidrule(lr){5-6} \cmidrule(lr){7-8} \cmidrule(lr){9-10} 
    \footnotesize ResNet-101 & \footnotesize DeepLabV2  &  44.0&60.4    &   43.4 & 59.6&   55.3& 76.0 &   57.9& 79.5& 36.0 & 72.8  \\
    \footnotesize MiT-B5 & \footnotesize DAFormer       &  47.5 & 60.8  &   47.1& 60.3 &   60.3& 77.2 &   67.6 &86.6& 47.6 & 78.1  \\
    \rowcolor{textcolor!30} \footnotesize EVA02-L & \footnotesize DAFormer     &  59.5 & 73.1  &   59.2& 72.7 &   70.3& 86.4 &   68.0& 83.5& 60.5 & 81.4  \\
    \bottomrule
  \end{tabular}}
  \vspace{0.01cm}
  \caption{\textbf{UDA performance on the GTA5 $\rightarrow$ Cityscapes} domain shift using different model architectures. "Source only" denotes training on $\gtavfull$, without any UDA methods, and "Oracle" training on $\cstrain$. "Abs." refers to the absolute mIoU on the $\csval$ dataset whereas "Rel." denotes the performance in \% relative to the oracle performance.}
  \label{tab:architecture_gta_cs}
\end{table}
 
Results for the three model architectures on the common synthetic-to-real domain shift  GTA5$\rightarrow$ Cityscapes are presented in \Cref{tab:architecture_gta_cs}.
We observe that equipping the simple DACS~\cite{tranheden2021dacs} method with the EVA02-L backbone causes a performance gain on Cityscapes of 10.0\% mIoU compared to the standard MiT-B5 encoder and also significantly raises performance relative to the oracle performance (supervised training on $\cstrain$) from 77.2\% to 86.4\%. We reason that similar to previous works~\cite{hummer2023vltseg, wei2024stronger, benigmim2024collaborating} the vision-language pre-training provides a stronger generalized backbone which better adapts to the target domain. However, the DAFormer~\cite{Hoyer2022daformer} approach does not benefit from the EVA02-L backbone and the performance remains at a level similar to that for the MiT-B5 backbone. This may be caused by the ImageNet feature distance (FD) loss which is designed to preserve the ImageNet pre-trained knowledge by minimizing the feature distance between the ImageNet and the synthetic object classes. We analyze the feature distance loss of the DAFormer~\cite{Hoyer2022daformer} training as plotted in \Cref{fig:fd_loss_comp}. The FD-loss is 5-10$\times$ higher for the EVA02-L backbone than for the MiT-B5 backbone and shows a different behavior at the beginning. 
This is reasonable since the loss is based on the ImageNet classes which do not align with the vision-language pre-trained EVA02-L backbone. The FD-loss is part of DAFormer~\cite{Hoyer2022daformer}, HRDA~\cite{hoyer2022hrda} and MIC~\cite{hoyer2023mic} and provides a performance increase of 3.5\% mIoU according to the original paper \cite{Hoyer2022daformer} but makes those methods hard to transfer to backbones which are not initialized with ImageNet pre-trained weights. The UDA methods AdaptSegNet~\cite{tsai2018learning} and ADVENT~\cite{Vu2019} do not
yield further gains to the source only performance with the EVA02-L backbone. These results indicate that the gain of a new backbone depends on the UDA method. 
\begin{table*}
\parbox[b]{.48\linewidth}{
\begin{tikzpicture}
  \begin{axis}[ 
  width=\linewidth,
  line width=1.0,
  height = 5cm, 
  grid=major, %
  tick label style={font=\normalsize},
  legend style={nodes={scale=0.8, transform shape}},
  label style={font=\normalsize},
  xlabel={Iterations},
ylabel={Feature Distance Loss},
   y tick label style={
    /pgf/number format/.cd,
    fixed,
    fixed zerofill,
 },
legend style={at={(1,1)}, anchor=north east,  draw=none, fill=none},
  ]
    \addplot[tu21] coordinates
      {(50, 0.64) (1000,0.4793) (2000,0.4736) (3000,0.4672) (4050,0.4699) (5000,0.4586)(6000,0.4400) (7000,0.4507) (8050, 0.4419) (9000, 0.4444) (10000, 0.4437) (11000, 0.4313) (12050, 0.4215) (13000, 0.4459) (14000, 0.4259) (15000, 0.4305) (16050, 0.4411) (17050, 0.4310) (18000, 0.4213) (19000, 0.4353) (20050, 0.4139) (21000, 0.4224) (22000, 0.4251) (23000, 0.4195) (24050, 0.4071) (25000, 0.4225) (26000, 0.4090) (27000, 0.4120) (28050, 0.4014) (29000, 0.4072) (30000, 0.4031) (31000, 0.4084) (32050, 0.4038) (33000, 0.3912) (34000, 0.4070) (35000, 0.4084) (36050, 0.3990) (37000, 0.4003) (38000, 0.4015) (39000, 0.3922) (39950, 0.4114)};
      \addlegendentry{EVA02-L}

      \addplot[tu31] coordinates
      {(50, 0.3247) (1000,0.2659) (2000,0.2840) (3000,0.2923) (4050,0.2709) (5000,0.2835)(6000,0.2869) (7000,0.2891) (8050, 0.2927) (9000, 0.2834) (10000, 0.2938) (11000, 0.2853) (12050, 0.2901) (13000, 0.2701) (14000, 0.2678) (15000, 0.2802) (16050, 0.2603) (17050, 0.2622) (18000, 0.2699) (19000, 0.2594) (20050, 0.2623) (21000, 0.2661) (22000, 0.2581) (23000, 0.2524) (24050, 0.2632) (25000, 0.2502) (26000, 0.2548) (27000, 0.2515) (28050, 0.2559) (29000, 0.2397) (30000, 0.2483) (31000, 0.2403) (32050, 0.2418) (33000, 0.2542) (34000, 0.2599) (35000, 0.2463) (36050, 0.2418) (37000, 0.2436) (38000, 0.2460) (39000, 0.2474) (39950, 0.2469)};
      \addlegendentry{ResNet-101}

       \addplot[tu41] coordinates
      {(50, 0.067) (1000,0.1134) (2000,0.1097) (3000,0.1187) (4050,0.1080) (5000,0.1125) (6000,0.1034) (7000,0.1031) (8050, 0.1114) (9000, 0.1070) (10000, 0.1005) (11000, 0.1016) (12050, 0.1078) (13050, 0.1031) (14000, 0.0969) (15050, 0.1046) (16050, 0.0949) (17050, 0.0981) (18000, 0.0932) (19000, 0.0966) (20050, 0.0913) (21000, 0.0902) (22000, 0.0875) (23000, 0.0895) (24050, 0.0923) (25000, 0.0881) (26000, 0.0887) (27000, 0.0894) (28050, 0.0887) (29000, 0.0873) (30000, 0.0878) (31000, 0.0823) (32050, 0.0846) (33000, 0.0852) (34050, 0.0888) (35000, 0.0859) (36100, 0.0795) (37000, 0.0841) (38000, 0.0828) (39000, 0.0849) (39950, 0.0848)};
      \addlegendentry{MiT-B5}
  \end{axis}
\end{tikzpicture}
\captionof{figure}{\textbf{Feature Distance Loss} over the 40k training iterations of the DAFormer~\cite{Hoyer2022daformer} adaptation with three different backbones.}
   
  \label{fig:fd_loss_comp}

}
\hfill
\vspace{-5mm}
\parbox[b]{.48\linewidth}{
\setlength{\tabcolsep}{1pt}
\scriptsize
\centering
  \footnotesize
  \resizebox{0.48\columnwidth}{!}{
  \begin{tabular}{c|c|ccc}
    \toprule
     \textbf{Encoder}&\textbf{Method} & \textbf{GTA5 $\rightarrow$ CS}    & \textbf{SYN $\rightarrow$ CS}      & \textbf{CS $\rightarrow$ ACDC }   \\
    \midrule    
    \multirow{7}{*}{ResNet-101}& AdaptSegNet~\cite{tsai2018learning}     &42.4 &46.7&33.4* \\
    &ADVENT~\cite{Vu2019}     &45.5 &48.0& 32.7* \\
    &DACS~\cite{tranheden2021dacs}  & 52.1 & 54.8  & - \\
    &DAFormer~\cite{Hoyer2022daformer}&56.0&-&-\\
    &SePiCo~\cite{xie2023sepico} &61.0 &66.5 &- \\
    &HRDA~\cite{hoyer2022hrda} &  63.0 & 69.2 &57.6* \\
    &MIC~\cite{hoyer2023mic} &  64.2 & 70.7 & 60.4* \\
    \hline
    \multirow{4}{*}{MiT-B5}&DAFormer~\cite{Hoyer2022daformer}   & 68.3 & 67.4   & 55.4*  \\
    &SePiCo~\cite{xie2023sepico} & 70.3& 71.4&59.1*\\
    &HRDA~\cite{hoyer2022hrda}  & 73.8 & 72.4 & 68.0*\\
    &MIC~\cite{hoyer2023mic}  & \textbf{75.8} & \textbf{74.0} & 70.4* \\
    \hline
    \rowcolor{textcolor!30} & DACS~\cite{tranheden2021dacs}     & 70.3 & 72.3   & \textbf{72.0}  \\
    \rowcolor{textcolor!30} \multirow{-2}{*}{EVA02-L}& DAFormer~\cite{Hoyer2022daformer}     & 68.0  & 64.7& 68.0 \\
    \bottomrule
  \end{tabular}}
  \vspace{0.1mm}
  \caption{\textbf{Comparison with state-of-the-art UDA methods.} * marks the performance on the ACDC test set $\acdctest$. All values taken from respective papers except EVA02-L.}
\label{tab:uda_sota} }
\end{table*}

In \Cref{tab:uda_sota} we compare the performance of DACS~\cite{tranheden2021dacs} and DAFormer \cite{Hoyer2022daformer} with the newly employed EVA02-L backbone to the published state-of-the-art performances of other approaches using a ResNet-101 and MiT-B5 backbone. We observe that the performance with the DACS method performs similar to recent works like SePiCo~\cite{xie2023sepico} for both GTA5 and SYNTHIA as the source domain. However, compared to MIC~\cite{hoyer2023mic} the performance is less which may be also caused by the lower resolution. Notably, the performance of DAFormer \cite{Hoyer2022daformer} reduces for SYNTHIA$\rightarrow$Cityscapes with the EVA02-L backbone but increases significantly for Cityscapes$\rightarrow$ACDC. We further evaluated on a pure adverse weather domain shift, using the ACDC~\cite{sakaridis2021acdc} reference images under normal weather conditions $\acdcnormal$ and their counterparts taken under adverse weather conditions e.g. $\acdcfog$ etc. As shown in \Cref{tab:acdc_clear_weather} we both adapted the model solely to the respective sub-domains and also to $\acdctrain$ which contains all sub-domains. First, we can see that the ranking of the different UDA methods differs to the ranking on the GTA5$\rightarrow$Cityscapes benchmark. Both AdaptSegNet \cite{tsai2018learning} and ADVENT \cite{Vu2019} perform better with the EVA02-L encoder on the ACDC shift than DAFormer. This may be related to the FD-loss of DAFormer. In contrast, the state-of-the-art UDA method MIC~\cite{hoyer2023mic} with the MiT-B5 backbone also shows the best performance on the ACDC shifts, mostly with a significant margin of up to 8.6\% mIoU on the rain dataset and 5.8\% difference for the mean. For certain cases we observe that UDA methods which perform better on the synthetic-to-real benchmark may perform worse on this benchmark. The DACS method with the EVA02-L backbone performs 5.6\% mIoU worse on the synthetic-to-real benchmark than MIC while being better by 5.0\% in the mean across the adverse weather conditions.\addtolength{\tabcolsep}{5pt}
\begin{table}[ht]
  \centering
  \vspace{-0.2cm}
  \resizebox{0.9\columnwidth}{!}{
  \begin{tabular}{c|l||c||cccc|c||c}
    \toprule
                   & & & \multicolumn{4}{c|}{\textbf{Target Domain $\tgt$}} &       &  $\tgt$   \\ %
   \textbf{Enc.}& \textbf{UDA Method}  &GTA5$\rightarrow$CS &       $\acdcfog$     &       $\acdcrain$    &       $\acdcsnow$    &       $\acdcnight$   &   mean & $\acdctrain$  \\
    \midrule[0.3pt]
    \midrule[0.3pt]
   \multirow{5}{*}{\parbox[t]{2mm}{\rotatebox[origin=c]{90}{ResNet-101}}}& Source Only & 36.0 &       65.2   &       51.9   &       52.8   &       32.4   &   50.6 &  50.0 \\
   &AdaptSegNet~\cite{tsai2018learning}& 44.0 &       64.4   &       55.6   &       54.5   &       35.8   &   52.6  & 51.6 \\
    &ADVENT~\cite{Vu2019}      & 43.4 &       64.1   &       53.0   &       54.1   &       34.1   &   51.3  & 52.2 \\
    &DACS~\cite{tranheden2021dacs}       & 55.3 &      68.0  &       \textbf{57.8}   &       \textbf{59.4}   &       37.8   &   \textbf{55.7}  & 55.6 \\
    &DAFormer~\cite{Hoyer2022daformer}  & \textbf{57.9} &   \textbf{68.2}   &  54.4 &  58.5 &       \textbf{38.5}   &   54.9  & \textbf{56.6} \\
    \midrule[0.3pt]
    \multirow{4}{*}{\parbox[t]{2mm}{\rotatebox[origin=c]{90}{MiT-B5}}}&Source Only&47.6&70.9   &       63.6   &       61.4   &       36.4   &   58.1&59.2\\
    &DAFormer~\cite{Hoyer2022daformer} & 67.6 &  73.5   &  59.2 &  64.5 &       47.1   &   61.1  & 64.2 \\
    &SePiCo~\cite{xie2023sepico}      & 67.3 &      76.7   &       62.7   &       63.7   &       47.6   &   62.7  & 66.2 \\
    &MIC~\cite{hoyer2023mic}         & \textbf{75.9} &     \textbf{79.6}   &       \textbf{71.3}   &      \textbf{69.1}    &       \textbf{53.5}   &   \textbf{68.4}  & \textbf{72.0} \\
    \midrule[0.3pt]
    \rowcolor{textcolor!30} & Source Only & 60.5 &       81.4   &       75.6   &       74.0   &       \textbf{59.0}   &  72.5  & 74.1 \\
    \rowcolor{textcolor!30}& AdaptSegNet~\cite{tsai2018learning} & 59.5 & 81.2 & \textbf{75.8}  & 74.3 & 54.7 & 71.5 & 72.2 \\
    \rowcolor{textcolor!30}& ADVENT~\cite{Vu2019} & 59.2 & 80.9 & 75.5  & 69.9 & 54.0 & 70.1 & 73.1 \\
    \rowcolor{textcolor!30}& DACS~\cite{tranheden2021dacs} & \textbf{70.3} & \textbf{82.6} & 75.6  & \textbf{77.7} & 57.6 & \textbf{73.4} & \textbf{75.6} \\
    \rowcolor{textcolor!30}\multirow{-5}{*}{\parbox[t]{2mm}{\rotatebox[origin=c]{90}{EVA02-L}}}&DAFormer~\cite{Hoyer2022daformer} & 68.0 &  78.3   & 70.4  & 74.2  &  53.9        &   69.2  & 70.3 \\
    \bottomrule
  \end{tabular}}
  \vspace{0.2cm}
  \caption{\textbf{UDA performance from $\acdcnormal$ to $\acdcfog$, $\acdcrain$, $\acdcsnow$ and $\acdcnight$.} The values for GTA5 $\rightarrow$ Cityscapes are given for comparison.}
  \label{tab:acdc_clear_weather}
\end{table}
However, the gain over the source only performance with the EVA02-L encoder on the pure ACDC shift is limited and only DACS \cite{tranheden2021dacs} provides a minor improvement. That shows that different UDA methods perform differently for different shifts and the 5.0\% gap to the performance of MIC \cite{hoyer2023mic} is mainly attributed to the pre-training of the encoder. In line with the results from~\cite{sakaridis2021acdc}, we observe that certain methods even show worse performances than a source-only trained model for certain domains. DAFormer~\cite{Hoyer2022daformer} with MiT-B5 adapted to the rain images reaches a 4.4\% mIoU lower performance. Also ADVENT~\cite{Vu2019} and AdaptSegNet~\cite{tsai2018learning} with a ResNet-101 encoder slightly reduce the performance on the fog domain. With the EVA02-L backbone all UDA methods except DACS \cite{tranheden2021dacs} reach a lower performance in average and the  DAFormer \cite{Hoyer2022daformer} performance drops by 3.3\% mIoU compared to source only. While this may be related to the FD-loss also AdaptSegNet \cite{tsai2018learning} and ADVENT \cite{Vu2019} undergo a clear performance drop of up to 2.4\% mIoU in average. The adaptation to $\acdctrain$ mostly leads to higher performance compared to the mean of adapting to single sub-domains especially for the MiT-B5 backbone, e.g. a gain of 3.6\% mIoU for MIC~\cite{hoyer2023mic}, and EVA02-L. For the ResNet-101 backbone the performance is mostly similar or even smaller. This might be caused by the different abilities of the vision transformer backbone who can utilize the larger amount of target data more effectively and benefit from the larger diversity of the target domain. 

\subsection{Domain generalization of UDA methods}
The domain generalization performance of UDA approaches to entirely unseen domains is rarely evaluated but highly relevant because the adaptation to e.g. a certain real target domain should intuitively also improve the generalization to other unseen target domains. For this reason, we evaluate the domain generalization performance across different backbones following the same protocol as pure DG approaches~\cite{hummer2023vltseg, niemeijer2024generalization, fahes2024simple} and compare it with state-of-the-art DG approaches similar to~\cite{piva2023empirical}.
\begin{table}[ht]
  \centering
  \footnotesize
  \vspace{2pt}
  \resizebox{0.9\linewidth}{!}{
  \begin{tabular}{c|r|c||ccc|c}
    \toprule
      \multicolumn{2}{l|}{$\src=\gtavfull$} & $\mathbf{\tgt}=\cstrain$    & \multicolumn{4}{c}{\textbf{Domain Generalization}}    \\ 
      \cmidrule[0.3pt]{3-7}
     \textbf{Enc.}&\textbf{UDA/DG Method} & \cellcolor{tu22} $ \csval$  &   \cellcolor{tu32} $\mvval$   &  \cellcolor{tu42}     $\bddval$    & \cellcolor{tu82}  $\acdcval$         &       \textbf{DG mean}   \\
    \midrule[0.3pt]
    \midrule[0.3pt]
    \rowcolor{gray!30}\cellcolor{white}& FAmix~\cite{fahes2024simple}  & 49.5 & 52.0 & 46.4 & \textbf{36.1} & \textbf{44.8} \\
    \rowcolor{gray!30} \cellcolor{white}&CLOUDS~\cite{benigmim2024collaborating}  & 55.7 & \textbf{59.0} & \textbf{49.3} & - & -\\ 
    \rowcolor{gray!30}\cellcolor{white}&VLTSeg~\cite{hummer2023vltseg}  &51.2 & 52.2 & 43.3 & - & - \\
    %\cline{2-7}
    &AdaptSegNet~\cite{tsai2018learning}   &  44.0   &  40.8    &       40.0    & 27.1        &   36.0        \\
    &ADVENT~\cite{Vu2019}        &  43.4   &  40.4    &       40.2    & 27.5         &  36.0        \\
    &DACS~\cite{tranheden2021dacs}         &  55.3   &  46.3    &       39.4    &  32.4        &   39.4      \\
    \multirow{-7}{*}{\parbox[t]{2mm}{\rotatebox[origin=c]{90}{ResNet-101}}}&DAFormer~\cite{Hoyer2022daformer}    &  \textbf{57.9}   &  48.4    &       41.6    &  35.0       &   41.7      \\
    \midrule[0.3pt]
    \rowcolor{gray!30} \cellcolor{white}& DGinStyle~\cite{jia2023dginstyle}  &58.6 & 62.5 & 52.3 & 46.1 & 53.6\\
    \rowcolor{gray!30} \cellcolor{white}&HRDA~\cite{hoyer2023domain}  &57.4 &61.2 &49.1 &44.0& 51.4\\ 
    \rowcolor{gray!30} \cellcolor{white}&CLOUDS~\cite{benigmim2024collaborating} &58.1 &62.3 & 53.8 & - & -\\ 
    \rowcolor{gray!30} \cellcolor{white} &DIDEX~\cite{niemeijer2024generalization}  &62.0 &63.0 &54.3 & 50.1& 55.8\\
    %\cline{2-7}
    &AdaptSegNet~\cite{tsai2018learning}   &   47.5  &   48.4   &        44.2   &     34.7     &  42.4        \\
    &ADVENT~\cite{Vu2019}        &   47.1  &   47.6   &        44.8   &    34.8    &  42.4         \\
    &DACS~\cite{tranheden2021dacs}         &   60.3  &   58.0   &        51.3   &   43.5       &  50.9        \\
    &DAFormer~\cite{Hoyer2022daformer}     &   67.6  &   58.6   &        52.2   &    45.5    &  52.1        \\
    &SePiCo~\cite{xie2023sepico} &67.3&60.0&52.3&47.8&53.4\\ 
    \multirow{-10}{*}{\parbox[t]{2mm}{\rotatebox[origin=c]{90}{MiT-B5}}} & MIC~\cite{hoyer2023mic} &\textbf{75.9} &\textbf{69.3} &\textbf{57.6} &\textbf{56.8} &\textbf{61.2}\\
    \midrule[0.3pt]
    \rowcolor{gray!30} \cellcolor{white}& VLTSeg~\cite{hummer2023vltseg}   &65.6 &66.5 &58.4 & 62.6 &62.5\\
    \rowcolor{gray!30} \cellcolor{white}& Rein~\cite{wei2024stronger}  &65.3 &66.1 &60.4 & - &-\\
    %\cline{2-7}
    \rowcolor{textcolor!30} \cellcolor{white}& AdaptSegNet~\cite{tsai2018learning}   &  59.5   &  63.1    &       56.0    &    54.3     &  57.8     \\
    \rowcolor{textcolor!30} \cellcolor{white}& ADVENT~\cite{Vu2019}        &  59.2   &  62.8    &       57.4    &    54.9     &     58.4     \\
    \rowcolor{textcolor!30} \cellcolor{white}& DACS~\cite{tranheden2021dacs}        &  \textbf{70.3}   &  \textbf{68.2}   &       \textbf{61.2}   &     \textbf{64.4}    &    \textbf{64.6}       \\
    \rowcolor{textcolor!30} \cellcolor{white} \multirow{-6}{*}{ \parbox[t]{2mm}{\rotatebox[origin=c]{90}{EVA02-L}}} & DAFormer~\cite{Hoyer2022daformer}    &  68.0   &  64.8    &       58.1    &     61.6    &      61.5     \\
    \bottomrule
  \end{tabular}}
  %\vspace{0.2cm}
   \caption{\textbf{Domain generalization (DG) and UDA performances} on various real datasets of GTA5$\rightarrow$Cityscapes UDA models and DG methods. The DG mean is calculated across Mapillary, BDD and ACDC. \colorbox{gray!30}{gray} marks domain generalization methods which were trained on GTA5 without any adaptation. For these, values are taken from the respective publications.} 
  \label{tab:uda_dg_performance}
\end{table}

The results are shown in \Cref{tab:uda_dg_performance}. Combining a simple UDA method like DACS~\cite{tranheden2021dacs} with EVA02-L improves the DG performance by 25.2\% over the ResNet-101 and by 13.7\% mIoU over the MiT-B5 backbone which highlights the strong generalization capabilities of vision-language pre-trained backbones. It also further improves DG performance compared to pure domain generalization methods and outperforms VLTSeg \cite{hummer2023vltseg} by 2.1\% mIoU in average across Mapillary Vistas, BDD100K and ACDC. That confirms the observation from Piva \etal~\cite{piva2023empirical} that UDA methods can provide a better generalization than DG methods. Intuitively, this can be expected since the adaptation to real images should also increase the performance on other unseen real domains since there are basic patterns which can be learned from the unlabeled real domain. The observation for the MiT-B5 backbone is similar. Recent DG methods perform similarly or outperform the generalization of several UDA methods with this backbone but cannot compete with the recent UDA approach MIC~\cite{hoyer2023mic}. DIDEX~\cite{niemeijer2024generalization} as a recent DG method outperforms both DACS~\cite{tranheden2021dacs} and DAFormer~\cite{Hoyer2022daformer} by 4.9\% and 3.7\% mIoU in the DG mean respectively. However, MIC~\cite{hoyer2023mic} performs 5.4\% mIoU better in the DG mean. In contrast, for the ResNet-101 backbone, CLOUDS~\cite{benigmim2024collaborating} outperforms DAFormer with its generalization by a large margin of 7.7\% and 10.6\% on BDD100K and Mapillary respectively. This is not a contradiction to the results of Piva \etal~\cite{piva2023empirical} since DG methods made a significant progress recently by e.g. using foundation models like CLOUDS \cite{benigmim2024collaborating} which enabled them to surpass the generalization of UDA methods. Notably, we observe that the UDA target domain performance on Cityscapes does not necessarily translate into a similar generalization performance. While we can see a clear performance gain of DAFormer over DACS with the MiT-B5 backbone on Cityscapes of over 7\% mIoU, the performance gap is reduced to 1.2\% mIoU on the DG mean. For other backbones we make similar observations, like a Cityscapes ResNet-101 performance difference of 11.3\% mIoU between AdaptSegNet and DACS but only 3.4\% mIoU difference on the DG mean. This may be related to a method-dependent overfitting of the UDA methods to the target domain which hampers the generalization of UDA methods.  
\addtolength{\tabcolsep}{-5pt}   

\begin{table}[ht]
  \centering
  \vspace{-0.0cm}
  \resizebox{0.9\linewidth}{!}{
  \begin{tabular}{c|l|c||ccc|c}
    \toprule
         \multicolumn{2}{l|}{$\src=\acdcnormal$}& $\mathcal{D}^T=\acdctrain$     & \multicolumn{4}{c}{\textbf{Domain Generalization}}    \\ \cmidrule[0.3pt]{3-7}
    \textbf{Encoder}&\textbf{UDA Method}  & \cellcolor{tu82} $\acdcval$ &\cellcolor{tu22} $\csval$      &\cellcolor{tu32}  $\mvval$   &       \cellcolor{tu42}  $\bddval$    &       \textbf{DG mean}    \\
    \midrule[0.3pt]
    \midrule[0.3pt]
    \multirow{4}{*}{ResNet-101}&AdaptSegNet~\cite{tsai2018learning} &  51.6 &  52.5  &  \textbf{52.0}   &       42.5   &       49.0    \\
    &ADVENT~\cite{Vu2019}      &  52.2   & 53.1 &  51.6   &       42.9   &       49.2            \\
    &DACS~\cite{tranheden2021dacs}     &  55.6&       55.0   &  48.9   &       42.1   &       48.7      \\
    &DAFormer~\cite{Hoyer2022daformer}  & \textbf{56.6} &       \textbf{56.3}  & 51.4   & \textbf{44.0}& \textbf{50.5}     \\
    \hline
    \multirow{3}{*}{MiT-B5}&DAFormer~\cite{Hoyer2022daformer} & 64.2 &       65.5 &   59.1   & 49.0 & 57.9       \\
    &SePiCo~\cite{xie2023sepico}      &  66.2  &       64.7   &  58.6   &       49.7   &       57.7   \\
    &MIC~\cite{hoyer2023mic}        &  \textbf{72.0}  &       \textbf{70.0}  &  \textbf{64.7}   &       \textbf{56.4}   &       \textbf{63.7}    \\
    \hline
    \rowcolor{textcolor!30} &AdaptSegNet~\cite{tsai2018learning}  &  72.2&75.4&68.5&\textbf{61.6}& 68.5 \\
    \rowcolor{textcolor!30} &ADVENT~\cite{Vu2019}  & 73.1 &75.3&68.2&61.0&68.2 \\
    \rowcolor{textcolor!30}&DACS~\cite{tranheden2021dacs}   &   \textbf{75.6}     &  \textbf{75.8}   &   \textbf{68.9}      & \textbf{61.6} & \textbf{68.8}  \\
    \rowcolor{textcolor!30} \multirow{-4}{*}{EVA02-L} &DAFormer~\cite{Hoyer2022daformer}  & 70.3 &  72.3   &  66.5  &  55.7 & 64.8     \\
    \bottomrule
  \end{tabular}}
  \vspace{0.2cm}
  \caption{\textbf{Domain generalization and UDA performances} for adaptation from ACDC clear weather reference images $\acdcnormal$ to all adverse ACDC conditions. DG Mean is calculated over Mapillary, BDD100K and Cityscapes.}
  \label{tab:dg_acdc_clean}
  \vspace{-0.2cm}
\end{table}
We also evaluated the DG performance on the pure adverse weather shift of the ACDC dataset. We can observe from \Cref{tab:dg_acdc_clean} that similar to the synthetic-to-real shift a higher performance in the target domain not necessarily causes a higher domain generalization performance. DACS~\cite{tranheden2021dacs} with ResNet-101 performs 4\% mIoU better on the adverse ACDC weather domains compared to AdaptSegNet~\cite{tsai2018learning} but 0.3\% worse on the DG mean. SePiCo~\cite{xie2023sepico} has a 2\% mIoU higher performance on the target domain than DAFormer~\cite{Hoyer2022daformer} but performs slightly worse in the DG mean. MIC~\cite{hoyer2023mic} shows its strong generalization capabilities also in this pure real-to-real benchmark and outperforms DAFormer by a clear margin of 5.8\% mIoU. Surprisingly, all four methods AdaptSegNet \cite{tsai2018learning}, ADVENT \cite{Vu2019}, DACS \cite{tranheden2021dacs} and DAFormer \cite{Hoyer2022daformer} with the EVA02-L backbone outperform MIC not only for the generalization performance by up to 5.1\% mIoU but also on the target domain by up to 3.6\% mIoU. This may be caused by the smaller target dataset compared to the adaptation to Cityscapes which increases the influence of the pre-trained representations of the EVA02-L backbone. It also shows how different UDA methods can perform on different domain shifts with different encoders since the DG ranking for GTA5$\rightarrow$Cityscapes is different.

\subsection{Discussion}
We did not apply any changes to the UDA methods like e.g. changing hyperparameters, a different resolution or disabling the FD-loss. This may have improved the performance of the UDA methods. However, in contrast to Englert \etal~\cite{englert2024exploring} our aim was to evaluate the UDA methods without any changes to assess how well they transfer to a different domain shift and a new encoder architecture with vision-language initialization. Modifying existing UDA methods may not be trivial because removing or adapting certain components will likely influence the performance and behavior. 

\section{Conclusion}
We equipped existing UDA methods with a state-of-the-art vision-language pre-trained encoder and studied the target performance and the generalization to unseen domains. The results demonstrate the potential of vision-language pre-training for UDA by reaching a competitive target domain performance with a simple UDA method. They also indicate strong generalization capabilities for both established benchmarks and a pure $\rightarrow$ adverse weather condition domain shift based on ACDC. We show that recent state-of-the-art UDA methods rely on a loss function which cannot be directly used for the vision-language pre-trained encoder. Our results indicate that similar to domain generalization new UDA methods are required to fully exploit the potential of vision-language models for UDA. Our domain generalization evaluations showed two novel findings. First, the target domain performance is not necessarily an indicator for their generalization capabilities and second, that recent, pure DG methods are performing in parts similarly or even superior than UDA methods.     

\bibliography{egbib}
\end{document}